\pdfoutput=1

\documentclass[11pt]{article}

\usepackage{acl}

\usepackage{times}
\usepackage{latexsym}
\usepackage{enumitem}
\usepackage{xspace}
\usepackage{booktabs}
\usepackage{multirow}
\usepackage{amsmath}
\usepackage{siunitx}
\usepackage[T1]{fontenc}

\usepackage[utf8]{inputenc}

\usepackage{microtype}

\usepackage{inconsolata}

\usepackage{graphicx}
\usepackage{amsmath}
\usepackage{amssymb}

\usepackage{url}

\newcommand{\knn}{$k$NN-LMs\xspace}

\title{Great Memory, Shallow Reasoning: Limits of \knn}

\author{Shangyi Geng \quad Wenting Zhao \quad Alexander M Rush\\
Cornell University\\
\texttt{\{sg2323, wz346, arush\}@cornell.edu}\\
}

\begin{document}

\maketitle

\begin{abstract}
$K$-nearest neighbor language models (\knn), which integrate retrieval with next-word prediction, have demonstrated strong performance in language modeling as well as downstream NLP benchmarks. These results have led researchers to argue that models trained on poor quality or outdated data could perform well by employing a $k$NN extension that has access to a higher-quality datastore. In this work, we ask whether this improved ability to recall information really translates into downstream abilities. We extensively evaluate \knn on a diverse set of tasks, ranging from sentiment classification and commonsense reasoning to multi-hop reasoning. Results show that \knn excel at \emph{memory}-intensive tasks, where utilizing the patterns in the input is sufficient for determining the output, but struggle with \emph{reasoning} tasks that require integrating multiple pieces of information to derive new knowledge. We further demonstrate through oracle experiments and qualitative analysis that even with perfect retrieval, \knn still fail to determine the correct answers, placing an upper bound on their reasoning performance. Code and datastores are released at \url{https://github.com/GSYfate/knnlm-limits/}.
\end{abstract}

\section{Introduction}
\label{sec:introduction}
\begin{figure*}[t]
    \centering
    \includegraphics[width=0.7\textwidth]{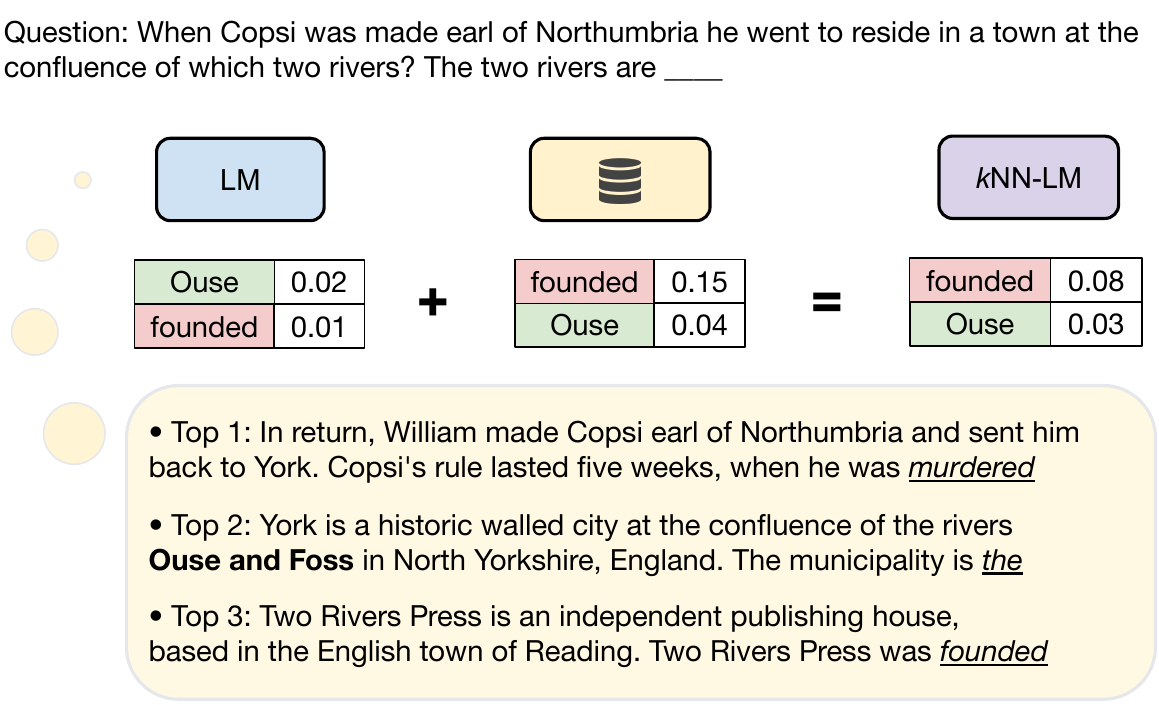}
    \caption{
    In this multi-hop question answering (QA) example, the LM is very uncertain about the next word and could benefit from retrieval. The $k$NN approach finds several document, both irrelevant and relevant, that may help. However, two issues occur: first, an irrelevant document increases the probability of a random wrong answer; second, even though a relevant document has been found, it may not upweight the actual answer (Ouse). We study how these issues may impact task performance as compared to perplexity.}
    \label{fig:reasoning}
\end{figure*}
A foundational property of pretrained language modeling~\citep{peters-etal-2018-deep, devlin-etal-2019-bert} has been that improvements to the perplexity of the model lead to improvements on downstream tasks. This property is central to the scaling of large language models (LLMs) where researchers focus nearly exclusively on perplexity as a proxy metric for improved general purpose abilities~\citep{kaplan2020scaling}. In recent years, this research has centered primarily on high-quality text data at greater and greater quantities as the limiting component for producing better language models~\citep{hoffmann2022training}.

This increasing need for data to train language models has led to significant challenges. On one hand, including as much high-quality data as possible results in improved downstream performance. On the other hand, this data is often protected by licenses or copyright, which means training on such data brings legal issues. 
For example, the recent high-profile lawsuit from the New York Times notes the clear use of their data in OpenAI models~\citep{grynbaum2023times}. 

It would be ideal to circumvent this issue entirely with alternative approaches. If a model could be trained on lower-quality data but adapted to perform well on real tasks, it might provide a technical workaround. Non-parametric Language Models (NPLMs), such as \knn, have emerged as a promising approach in this space~\citep{Khandelwal2020Generalization}. \knn extend neural LMs by linearly interpolating with simple k-nearest neighbor LMs. 
This approach can improve language modeling with its memory over a massive collection of texts, usually referred to as a datastore. \citet{khandelwal2021nearest} and \citet{shi-etal-2022-nearest} validate that \knn achieve better performance on downstream tasks compared to standard LMs. The SILO model of \citet{min2024silo} applies this approach further by training a LM exclusively on license-permissive data, and using a non-parametric datastore to improve the models during inference.

In this work, we study the limits of how \knn can be used to improve LLMs. Specifically, we are interested in whether the improvements in perplexity seen with \knn are equivalent to other improvements in LM ability, or if improvements in non-parametric memory are orthogonal to standard language modeling. This question relates to debates about whether memory is separable from other language abilities and how they interact in NLP benchmarks.

To study this question, we implement large-scale \knn on top of modern open LLMs with two datastores in different domains. We replicate past results that demonstrate significant decreases in perplexity across domains. This perplexity decrease transfers to similar benefits in task accuracy across several NLP benchmarks. These benchmarks are rather simple, where recognizing the patterns in the input and matching them with the patterns in memory is sufficient for determining the output. We refer to these as \textit{memory}-based tasks.

However, we see a different story when applying these models to tasks that require significant \textit{reasoning} ability. These tasks often require integrating multiple pieces of information to derive new knowledge. In our experiments, the use of \knn does not improve performance in reasoning, and in fact seems to hurt reasoning ability across tasks significantly. This behavior is robust and occurs even in domains that are explicitly targeted by the datastore used by the non-parametric model. These experiments lead us to conclude that while \knn may be useful in settings where data is constrained, they should not be seen as a remedy for low-quality training data, and that perplexity scores should not be seen as a corollary for LM ability outside of parametric training settings. 


\section{Related Work}
\label{related}
\paragraph{Retrieval Models}
Although Large Language Models (LLMs) achieve superhuman performance on a wide range of natural language processing tasks, they often produce hallucinations, struggle with integrating new knowledge, and expose private information present in the training data. Recently, research interest has shifted towards retrieval-based LMs, which combine a parametric neural model and a non-parametric external datastore \citep{guu2020retrieval, karpukhin-etal-2020-dense}. These retrieval-based LMs naturally incorporate new knowledge, enhance the factuality of generated texts, and reduce privacy concerns \citep{asai2024reliable}. Furthermore, studies \citep{borgeaud2022improving} have demonstrated that employing retrieval augmentation during large-scale pre-training can outperform standard LMs while requiring fewer parameters.

Among retrieval-based LMs, \knn \citep{Khandelwal2020Generalization} emerge as a popular choice \citep{min2024silo}. Unlike other retrieval models that encode and retrieve documents, \knn encode and retrieve tokens. At every token, \knn search for the $k$ most similar tokens from the datastore based on contextualized token embeddings, which are then turned into a next-token distribution. \knn linearly interpolate the retrieved $k$NN distribution with the output of a base LM. They do not require additional training but introduce computational and memory overhead.

\paragraph{Reasoning Retrieval.} Little research has been conducted on constructing retrieval models for reasoning tasks. Leandojo \citep{yang2023leandojo} investigates the use of retrieval-based LMs to assist with theorem proving, and \citet{levonian2023retrieval} experiment with retrieving content from mathematical textbooks to generate responses to student questions. In our study, we create a reasoning-specific datastore to assist LMs in performing reasoning-intensive tasks.

\paragraph{Evaluation of \knn.} 
While \knn excel at language modeling and have demonstrated enhanced performance in machine translation \citep{khandelwal2021nearest} and simple NLP tasks \citep{shi-etal-2022-nearest}, the question of whether they are thoughtful reasoners remains open. \citet{wang-etal-2023-knn} demonstrate that \knn struggle with open-ended text generation as they only provide benefits for a narrow set of token predictions and produce less reliable predictions when generating longer text. \citet{behnamghader-etal-2023-retriever} showed that when retrieval is conducted based on the similarity between queries and statements, \knn often fail to identify statements critical for reasoning. Even when these crucial statements are retrieved, it is challenging for \knn to effectively leverage them to infer new knowledge. These studies, however, are limited to a narrow set of tasks. Our work seeks to provide a comprehensive evaluation of the reasoning capabilities of \knn and provides an extensive analysis of the sources of their failures.

\section{\textit{k}-Nearest Neighbor Large Language Models}
Non-parametric language models are variants of standard language models that give the model the ability to utilize an additional datastore $\cal D$ during inference to determine the next word prediction, $p(x_{t+1} | x_{1\ldots t}; {\cal D} )$. This datastore may be part of the original training data, data for adaptation to a new domain, or be used to incorporate continual updates or protected data. As these datastores are typically quite large, this process requires a retrieval component in the loop to find the sparse subset of the datastore that can best inform the current prediction. Several popular approaches exist including DPR \citep{karpukhin-etal-2020-dense} and REALM \citep{guu2020retrieval}.

In this work, we focus on \knn due to their popularity as an approach to directly improve LM perplexity on fixed models without a need for retraining. As noted in the intro, this approach has also been put forward as a method for circumventing the need for high-quality licensed training data in LLMs. Formally \knn are defined as 
\begin{multline}
p(x_{1:T} ; \mathcal{D}) = \prod_{t} p(x_{t+1} \mid x_{1:t} ;\mathcal{D}) \\
= \!\prod_{t} \left( \lambda p_{\text{$k$NN}}(x_{t+1} \!\mid \! x_{1:t}; \mathcal{D}) \right.\!
\left. \!+\! (1-\lambda) p(x_{t+1} \!\mid\! x_{1:t}) \right) \nonumber
\end{multline}

Let $(k_i,v_i)$ be the $i$th (key, value) pair in $\mathcal{D}$, $f(\cdot)$ maps a token sequence to its contextual representation, and $d(\cdot)$ measures the distance between two vectors.
\begin{align*}
p_{\text{$k$NN}}&(x_{t+1} \mid x_{1:t}; \mathcal{D})\\ &\propto \sum_{(k_i,v_i) \in \mathcal{D}} \mathbf{1}_{x_{t+1}=v_i}
\times \exp(-d(k_i, f(x_{1:t}))).
\end{align*}


When using a Transformer language model, we define the distance metric $d(\cdot)$ as the squared \(\ell_2\) distance. To assemble the datastore we run the language model over all the documents to collect the necessary hidden states and corresponding next word. 

\paragraph{Experimental Setup.}
The hyperparameters include $\lambda$, $k$, and $\sigma$. $\lambda$ determines the weight of the datastore, and we consider $\lambda \in \{0.1, 0.2, 0.3\}$. Additionally, we retrieve $k \in \{1600, 2048\}$ neighbors and smooth the kNN distribution with a temperature $\sigma \in \{1, 3, 5, 10\}$.

For each inference model, we use Math and Wiki datastores for language modeling on the corresponding evaluation datasets: wikitext and math textbooks. Each datastore represents a specific domain, and we evaluate the performance of kNN-LM on a domain by measuring the perplexity of each evaluation dataset. We conduct a grid search to find the hyperparameters that yield the lowest PPL for each datastore. The optimal hyperparameters for each datastore are later applied across all downstream tasks in our experiments.

We provide eight demonstrations for GSM8K and three demonstrations for BBH. For the other datasets, we all perform zero-shot inference. We present full details of the experiments in the Appendix~\ref{sec:implementation}.

\paragraph{Inference and Retrieval Models.} We use Llama-2-7b \citep{touvron2023llama}, Llama-3-8B \citep{llama3modelcard}, and Mistral-7B~\citep{jiang2023mistral} as our inference models. For each inference model, we build the corresponding datastores. The keys are the 4096-dimensional hidden representations before the final MLP which predicts the token distribution at each generation step, produced by executing forward passes over the datastore corpora. For efficient similarity search, we create a FAISS index \citep{johnson2019billion} and search for nearest-neighbor tokens using Euclidean distance. Due to the scale of the datastores, we perform approximate search instead of exact search. We base our implementation on RetoMaton \citep{alon2022neuro}.

\begin{table}[t]\centering
    \begin{tabular}{lccc}
    \toprule
        $\cal D$ & Text Size & Tokens & Mem \\ \midrule
        Wiki & 2.2GB& 610M & 44G \\
        Math & 0.6GB& 200M & 15G\\ \bottomrule
    \end{tabular}
    \caption{Overview of the two datastores. Tokens are produced by Llama2 tokenizers. Mem is the memory size of the datastore.}
    \label{tab:overview}
\end{table}
\begin{table}[t]\centering
\begin{tabular}{lcc}
    \toprule
            & \multicolumn{2}{c}{LM Performance} \\
  Model         & Wiki  & Math \\ \midrule
Llama2-7b         & 10.63           & 7.90      \\
+Wiki     & \textbf{9.74}    & 8.75      \\
+Math & 11.33        &   \textbf{7.23} \\ \midrule
Llama-3-8b         & 9.70             & 5.36     \\ 
+Wiki     & \textbf{9.32}    & 6.03     \\
+Math & 10.37        & \textbf{5.22} \\ \midrule
Mistral-7B       & 9.72        & 5.64      \\
+Wiki     & \textbf{9.29}    & 6.41      \\ 
+Math & 10.49           & \textbf{5.59} \\\bottomrule
\end{tabular}
\caption{Perplexity comparison. Rows vary the datastore $\cal D$ used. Columns represent different held-out test sets. Lower numbers indicate better performance.}
\label{tab:perplexity}
\end{table}

\section{\knn Help In-Domain Perplexity}
To explore how different sources of external knowledge impact downstream task performance, we experiment with two datastores. First, we follow the choice made by \citet{shi-etal-2022-nearest}, where they identify heterogeneous data sources that are broadly relevant to common downstream NLP tasks. In particular, they mix Wikitext103 \citep{merity2017pointer}, with other sources including the English portion of Amazon Review~\citep{he2016ups}, and CC-NEWS~\citep{Hamborg2017} and IMDB~\citep{maas-EtAl:2011:ACL-HLT2011}. We call this datastore \emph{Wiki}. 


Then, we hypothesize that the commonly explored corpora for building datastores do not contain relevant knowledge to assist with math reasoning tasks. To maximize the performance gain on these tasks, we construct a datastore comprising 3.94K mathematical textbooks, sourced from \cite{wang2023generative}. These textbooks contain both theorems and practice questions, from which humans acquire mathematical knowledge. This datastore consists of 200M tokens. We will refer to this datastore as \emph{Math}. We summarize the statistics of each datastore in Table~\ref{tab:overview}.


We begin by validating past results of \knn on language modeling. We present results in Table~\ref{tab:perplexity}. To facilitate meaningful comparisons between models with different tokenizers and vocabulary sizes, we report word-level perplexities. These results show that having access to a non-parametric datastore leads to lower perplexity compared to using a standalone LM across all datasets. This improvement in perplexity is observed when the corpus used to construct the datastore and the one used for inference share the same data source. For instance, since the training split of Wikitext103 is in Wiki, the LM+Wiki setting achieves the lowest perplexity on Wikitext103's validation set. Utilizing the other datastore results in performance worse than that of the standalone LM. 

\section{\knn Can Help Memory-Intensive Tasks}

\begin{table*}[t]
\centering
\begin{tabular}{lccccccccc}
\toprule
          & RTE   & RT    & CB    & Yahoo & CR    & AGN   & HYP   & MR    & SST2  \\ \midrule
Llama2-7B        & 66.06 & 79.74 & 50.00 & \textbf{59.37} & 74.55 & 81.30 & \textbf{64.15} & 83.10 & 84.02 \\
+Wiki & \textbf{66.43} & 79.46 &\textbf{51.79} & 58.83 & \textbf{76.95} & 81.46 & \textbf{64.15} & 82.85 & \textbf{84.68} \\
+Math     & 65.70  & \textbf{82.55}      &  \textbf{51.79}    &  59.10     &  73.70     &\textbf{ 81.79 }   & 50.39     & 82.90     & 84.62      \\	\midrule	
Llama3-8B        & \textbf{70.76} & \textbf{79.46} & 64.29 & 58.87 & 79.10 & 79.17 & 59.30 & 83.80 & 86.54 \\
+Wiki & {61.37} & 79.55& \textbf{71.43} & \textbf{58.93} & \textbf{80.45} & 79.33 & 59.30 & 83.50 & 87.04 \\
+Math     & \textbf{70.76}  & 77.39      &  66.07     &  56.83    &  79.40    &\textbf{80.11}     &  59.30     & \textbf{84.30}      & \textbf{87.10}     \\	\midrule
Mistral-7B        & 76.17 & \textbf{75.32} & 71.43 & 56.63 & 81.90 & 73.57 & 56.59 & \textbf{79.35} & \textbf{81.82} \\
+Wiki & 76.17 & 75.05 & 67.86 & 56.63 & \textbf{82.15} & 73.55 & \textbf{56.78} & 79.30 & 81.77 \\
+Math     & 76.17  & 75.05      &  \textbf{75.00}     &  56.63     &  81.85     & \textbf{73.59}     &  \textbf{56.78}    & 79.10      & 81.77      \\	
\bottomrule
\end{tabular}
\caption{Accuracy comparison on various memory-intensive tasks.}
\label{tab:nlp}
\end{table*}

We begin by looking at a set of memory-intensive tasks, which we believe can be solved by pattern matching at scale without complex reasoning. We incorporate three types of tasks: sentiment classification, which aims to predict whether the sentiment of a text is positive or negative; textual entailment, which assesses the relationship between two sentences, determining if it constitutes entailment, contradiction, or neutrality; and topic classification, which involves identifying the main topic of a text. The datasets included for these tasks are as follows:
\begin{itemize}[leftmargin=*]
    \item For sentiment classification, we include SST-2 \citep{socher-etal-2013-recursive}, movie review (MR) \citep{mrdata}, customer review (CR) \citep{hu2004mining}, Rotten Tomatoes (RT), and a variant of hyperpartisan news detection (HYP) \citep{kiesel2019semeval}.
    \item For textual entailment, we use CommitmentBank (CB) \citep{de2019commitmentbank} and Recognizing Textual Entailment (RTE) \citep{dagan2010recognizing}.
    \item For topic classification, our datasets are AG News (AGN) \citep{zhang2015character} and Yahoo! Answers (Yahoo) \citep{zhang2015character}.
\end{itemize}

For classification and multiple-choice question-answering (QA) tasks, we utilize Domain Conditional Pointwise Mutual Information (DCPMI) \citep{holtzman-etal-2021-surface} to predict answers. We then calculate accuracy metrics to compare performance across different models. We measure the performance using F1 scores at the token level for text generation. Additionally, whenever feasible, we employ fuzzy verbalizers \citep{shi-etal-2022-nearest} to maximize the performance of \knn.

The results of these tasks are summarized in Table~\ref{tab:nlp}. On these tasks, \knn exhibit improved performance. Incorporating an external datastore outperforms a standalone LM on eight datasets while showing comparable performance on the remaining dataset. We further explain this performance gap through qualitative analysis in Appendix~\ref{sec:more-analysis}.

\section{\knn \textit{Hurt} Reasoning Performance}
\label{sec:evaluation}
For reasoning tasks, we consider three types: knowledge-intensive reasoning, which focuses on utilizing world knowledge for making (potential) multi-hop inferences; commonsense reasoning, which involves leveraging commonsense knowledge to understand social and physical interactions; and mathematical reasoning, which includes arithmetic, logical, and discrete reasoning abilities. The datasets selected for these categories are as follows:
\begin{itemize}[leftmargin=*]
    \item For knowledge-intensive reasoning, we explore Natural Questions (NQ) \citep{kwiatkowski2019natural}, HotpotQA \citep{yang-etal-2018-hotpotqa}, ARC Easy and Challenge \citep{allenai:arc}, OpenbookQA (OBQA) \citep{OpenBookQA2018}, and MMLU \citep{hendrycks2020measuring} to assess the model's ability to apply extensive world knowledge.
    \item For commonsense reasoning, we examine HellaSwag \citep{zellers-etal-2019-hellaswag} and Winogrande \citep{sakaguchi2021winogrande}, which test the model's understanding of social norms and physical laws.
    \item For mathematical reasoning, we utilize DROP \citep{dua-etal-2019-drop}, GSM8K \citep{cobbe2021training}, and Big Bench Hard (BBH) \citep{suzgun2022challenging} to evaluate the model's capacity for complex arithmetic, logical deductions, and handling of discrete concepts.
\end{itemize}

\begin{table*}[t]
\centering
\begin{tabular}{lcccccc}
\toprule
          & NQ & HotpotQA & Arc-Challenge & Arc-Easy & OBQA & MMLU \\ \midrule
Llama2-7B       & \textbf{23.18} & \textbf{22.72} & \textbf{41.81} & \textbf{57.49} & \textbf{57.00} & \textbf{39.22} \\
+Wiki           & 22.53 & 22.53 & 38.31 & 57.41 & 56.20 & 38.68 \\
+Math           & 21.14 & 21.26 & 41.04 & 56.82 & 56.20 & 38.53 \\ \midrule
Llama3-8B       & 23.64 & \textbf{25.14} & \textbf{44.88} & \textbf{58.83} & \textbf{55.80} & \textbf{42.67} \\
+Wiki           & \textbf{24.00} & 24.48 & 43.94 & 58.59 & 53.80 & 42.32 \\
+Math           & 23.04 & 24.63 & 43.26 & 58.59 & 54.60 & 42.46 \\ \midrule
Mistral-7B         & \textbf{20.63} & \textbf{20.96} & \textbf{46.42} & \textbf{60.94} & \textbf{58.80} & \textbf{41.91} \\
+Wiki           & 20.58 & 20.80 & 46.16 & 60.61 & 57.40 & 41.80 \\
+Math           & 20.56 & 20.48 & 46.08 & 60.77 & 57.80 & 41.55 \\\bottomrule
\end{tabular}
\caption{Performance comparison on datasets for knowledge-intensive reasoning tasks.}
\end{table*}
\label{tab:knowledge-reasoning}

\begin{table*}[t]
\centering
\begin{tabular}{lccccc}
\toprule
          & Winogrande & HellaSwag & DROP  & GSM8K & BBH   \\ \midrule
Llama2-7B      & 69.37 & \textbf{64.46} & \textbf{32.39} & \textbf{14.83} & 30.69 \\
+Wiki          & \textbf{70.32}          & 63.67          & 32.14          & 12.05          & \textbf{32.08} \\
+Math          & 68.98          & 63.54          & 32.31          & 13.48          & 30.82 \\ \midrule
Llama3-8B      & 73.95 & \textbf{65.99} & \textbf{45.55} & \textbf{45.72} & 39.67 \\
+Wiki          & 73.95          & 64.71         & 45.02         & 44.28          & 39.01 \\
+Math          & \textbf{74.19 }         & 65.15          & 45.54          & 45.63          & \textbf{39.92} \\ \midrule
Mistral        & 74.19 & \textbf{69.08} & \textbf{46.93} & 36.30 & \textbf{43.37} \\
+Wiki          & \textbf{74.66}         & 68.21          & 46.69          & 36.45          & 42.69 \\
+Math          & 73.64          & 68.11          & 46.38          & \textbf{36.60}        & 43.09 \\\bottomrule
\end{tabular}
\caption{Performance comparison on datasets for other reasoning tasks.}
\label{tab:other-reasoning}
\end{table*}

\begin{table}[t]
\small
    \centering
    \begin{tabular}{llSS}
    \toprule
        &           &  {Perplexity} & {Accuracy} \\ \midrule
        \multirow{2}{*}{OBQA}&LM & 255.76 & 55.80 \\
        &$k$NN-LM & 9.41 & 95.60\\   
        \multirow{2}{*}{NQ}&LM & 112.56 & 23.64 \\
        &$k$NN-LM & 8.91 & 46.40 \\
        \multirow{2}{*}{HotpotQA}&LM & 158.26 & 25.14 \\
        &$k$NN-LM & 8.15 & 49.85 \\
        \bottomrule
    \end{tabular}
    \caption{Results in an oracle setting where the \knn always include the correct answer as one of the $k$ nearest neighbors.}
    \label{tab:oracle-experiment}
\end{table}

\begin{table*}[t]
    \centering
    \small
    \begin{tabular}{p{10.5cm} p{2cm} p{2cm}}
        \toprule
        HotpotQA Example & Label & LM Pred   \\
        \midrule
          Which American character actor who starred on the television series ``Stargate SG-1'' (1997–2007) and appeared in ``Episode 8'' of ``Twin Peaks'' as a guest star? & \multirow{2}{*}{Don S. Davis} & \multirow{2}{*}{Don S. Davis} \\ \midrule
        Retrieved Context & Token & $k$NN-LM Pred \\ \midrule
        $\bullet$ After the first three seasons of Stargate SG-1 had been filmed on 16 mm film (although scenes involving visual effects had always been shot on 35 mm film for various technical reasons), ``Nemesis'' was the first episode filmed entirely on 35 mm film ... ``Nemesis'' was the last episode before actor & \multirow{4}{*}{Christopher} & \multirow{10}{*}{Michael Shanks} \\
        $\bullet$  
         ``200'' won the 2007 Constellation Award for Best Overall 2006 Science Fiction Film or Television Script, and was nominated for the 2007 Hugo Award for Best Dramatic Presentation, Short Form. The episode also marks the first time original SG-1 member& \multirow{4}{*}{Jack} &  \\
        $\bullet$  Season one regular cast members included Richard Dean Anderson, Amanda Tapping, & \multirow{2}{*}{Michael} &  \\
        \bottomrule
    \end{tabular}
    \caption{A multihop reasoning example from HotpotQA with predictions of the standard LM and \knn.}
    \label{tab:qualitative-hotpotqa}
\end{table*}

We present the results for knowledge-intensive tasks in Table~\ref{tab:knowledge-reasoning}. In stark contrast to the earlier findings, using a standalone LM consistently outperforms \knn on these tasks. Most surprisingly, on Natural Questions and HotpotQA, which consist of QA pairs constructed from Wikipedia documents, performance does not improve even though Wiki contains several million Wikipedia tokens. Retrieving from Wiki leads to a three-point decrease in performance.

Results for commonsense reasoning and mathematical reasoning tasks are shown in Table~\ref{tab:other-reasoning}. The standalone LM once again outperforms \knn models on four out of the five datasets. The most significant differences in performance occur on GSM8K.  Although incorporating an external data store results in a slight performance increase on Mistral, this does not demonstrate the effectiveness of \knn on GSM8K. Under Mistral's parameter settings,\knn has minimal changes on the predictions of the standalone LM, merely introducing some randomness.  Finally, although \knn do not improve GSM8K and Drop over standard LMs, we find that retrieving from Math improves over retrieving from Wiki.

\begin{table*}[t]
    \centering
    \small
    \begin{tabular}{p{11.8cm} c c}
    \toprule
        NQ Example & Label & LM Pred  \\ 
        \midrule
        who is the largest supermarket chain in the uk? & Tesco & Tesco \\ \midrule
        Retrieved context & Token & $k$NN-LM Pred\\ \midrule
        $\bullet$ The majority of stores will open as normal across the UK, however Sainsbury's advise shoppers to check details of when your local branch as some may close earlier than normal using the online store locator tool.(Image: Bloomberg) Supermarket giant & \multirow{3}{*}{Asda} &\multirow{7}{*}{Asda} \\
        $\bullet$ Along with Lidl, Aldi has eaten away at the market share of the Big Four supermarkets: &Tesco&\\
        $\bullet$ buy one, get one free (BOGOF) offers have been criticised for encouraging customers to purchase food items that are eventually thrown away; as part of its own campaign on food waste, supermarket retailer &\multirow{3}{*}{Morris}&\\
      
    \bottomrule
    \end{tabular}
    \caption{A knowledge-intensive reasoning example from Natural Questions with predictions of the standard LM and \knn.}
    \label{tab:qualitative-nq}
\end{table*}

\begin{table*}[t]
    \centering
    \small
    \begin{tabular}{p{9cm} p{2.5cm} p{2.5cm}}
    \toprule
        HotpotQA Example & Label & LM Pred  \\ 
        \midrule
       What type of plane is the four engine heavy bomber, first introduced in 1938 for the United States Army, which is hangared at Conroe North Houston Regional Airport? & American Boeing B-17 Flying Fortress & The B-17 Flying Fortress \\
        \midrule
        Retrieved context & Token & $k$NN-LM Pred\\
        \midrule
        $\bullet$ A famous symbol of the courage and sacrifices made by American bomber crews during World War II was revealed May 16 at the National Museum of the U.S. Air Force, Wright-Patterson Air Force Base, Ohio. The meticulously restored B-& \multirow{4}{*}{17}& \multirow{9}{*}{The B-25 Mitchell.}\\
        $\bullet$ As the Avenger made its way to the tower area, the wings began to fold up, a maneuver which enabled more of its kind to be loaded side by side into aircraft carriers. The queen of the event was the B- & \multirow{2}{*}{25} & \\
        $\bullet$ Spring is here, so why not hop a plane and grab some lunch? Even better if a World War II-era B- & \multirow{2}{*}{25} & \\
    \bottomrule
    \end{tabular}
    \caption{Example from HotpotQA showing the impact of high-frequency proper nouns in the corpus on \knn predictions retrieving from Wikipedia.}
    \label{tab:qualitative-high_freq}
\end{table*}

\begin{table*}[t]
    \centering
    \small
    \begin{tabular}{p {9cm} c c}
    \toprule
        Mathematical Reasoning Example & Label & LM Pred  \\ \midrule 
       I have three violins, three trombones, a flute, and four trumpets. How many musical instruments do I have? & \multirow{2}{*}{11} & \multirow{2}{*}{11} \\ \midrule
        Retrieved Context & Token  & $k$NN-LM Pred \\
        \midrule

        $\bullet$ In this example, the optimal route would be: 1 -> 3 -> 2 -> 4 -> 1, with a total completion time of  & 10 & \\
        $\bullet$ How many different passwords are there for his website system? How does this compare to the total number of strings of length & 10 & 10 \\
        $\bullet$ Using the TSP, the most efficient order in which to schedule these tasks would be: 2 -> 3 -> 1 -> 4 -> 2, with a total completion time of  & 14 &\\
        \bottomrule
        \end{tabular}
    \caption{A mathematical reasoning example from BBH requiring object counting with predictions of the standard LM and \knn.}
    \label{tab:qualitative-math}
\end{table*}

\section{Analysis}
\label{sec:analysis}
The results of this work show that \knn generally hurt reasoning of models, despite helping perplexity and other simpler tasks. In this section, we investigate the cause of this further. 

\paragraph{Qualitative Analysis.}
We conduct qualitative analysis to understand the failures of kNN-LMs better. In the qualitative analysis, we inspect examples of knowledge-intensive and mathematical reasoning datasets and show the retrieved tokens as well as the proceeding context. Through these examples, we find the following patterns that prevent kNN-LM from retrieving the correct token.

\begin{itemize}[leftmargin=*]
  \item \textbf{\knn struggle with multi-hop reasoning questions.} When the task requires extracting multiple pieces of sentences from the corpus and then combining the information to infer the answer, \knn often retrieve tokens that are contextually appropriate and relevant to part of the question, rather than the correct answer. As shown in Table~\ref{tab:qualitative-hotpotqa}, for the multi-hop reasoning question from HotpotQA, the model needs to identify an actor who both starred in Stargate SG-1 and guest-starred in Twin Peaks. While the required information is available in Wikipedia, it is distributed across two paragraphs. \knn retrieve only the actors from Stargate SG-1, failing to combine information from two sources to perform accurate multi-hop reasoning.
  \item \textbf{\knn are sensitive to the syntax but not the semantics of the question.} While $k$NN-LM retrieves the next token that fits the context, it cannot distinguish subtle semantic differences between different words in a sentence. As a result, when more than one word fits the context, it may not select the correct answer. Table~\ref{tab:qualitative-nq} demonstrates this issue with an example from the NQ dataset. Even though Asda is not the largest supermarket in the UK, due to the highly similar contexts of `supermarket giant' and `the largest supermarket, \knn ultimately assign a high probability to Asda and make a wrong prediction.
  \item \textbf{\knn tend to retrieve high-frequency entities in the corpus.} The entities are often proper nouns like person names and locations. If part of the answer overlaps with these high-frequency proper nouns, \knn will retrieve them and make wrong predictions, as shown in Table~\ref{tab:qualitative-high_freq} and Table~\ref{tab:qualitative-high_freq_2}.
  \item \textbf{\knn fail at mathematical reasoning tasks.} For instance, in the object counting task from the BBH dataset, even though kNN-LM understands the context that it needs to retrieve a number as the next token, it cannot solve the complex task of first identifying which objects are musical instruments and then counting them, as shown in Table~\ref{tab:qualitative-math}. 
\end{itemize}

\paragraph{Is the problem a failure of model weighting?} We investigate whether degraded reasoning capabilities of \knn stem from a failure in choosing a good weighting $\lambda$. This experiment aims to analyze \knn' behaviors when $\lambda$ is optimal for the downstream task. Specifically, we directly search for $\lambda$ that maximizes the log probabilities of a small set of labeled downstream task examples. We conduct this experiment on OpenbookQA and HotpotQA. We enumerate through retrieving $k\in \{16,32,64,128,256,512,1024,2048\}$ neighbors and setting temperature $\sigma \in \{1, 2, 5, 10\}$. We retrieve from Wiki. We initialize $\lambda$ at 0.5, and as the optimization proceeds, we find that smaller $\lambda$ values correlate with lower loss. Ultimately, we arrive at the minimum loss when $\lambda$ is close to 0. This process suggests that without any interpolation of the $k$NN distribution, the correct labels of the provided demonstrations receive the highest log probability. Therefore, OpenbookQA and HotpotQA are unlikely to benefit from having simple $k$NN access to Wiki.

\paragraph{Is the problem a failure of retrieval?} We investigate whether degraded reasoning capabilities of \knn stem from a failure in retrieval. We examine \knn' behaviors when retrieval is perfect. To achieve perfect retrieval, we include the correct answer among the $k$ nearest neighbors. Specifically, we construct a datastore for OpenbookQA, NQ, and HotpotQA, respectively, including their train and test examples. We then examine both perplexity and accuracy. The results, presented in Table~\ref{tab:oracle-experiment}, indicate that while \knn can significantly reduce the perplexity, the model does not always derive the correct answer, even when the correct answer is explicitly given as one of the $k$ neighbors. Therefore, the failure of reasoning cannot be fully attributed to the failure of retrieval. However, perfect retrieval does improve LM by a large margin, suggesting that better retrieval is beneficial. Currently, retrieval is performed by finding similar hidden representations. A training-based approach such as RAG \citep{lewis2020retrieval} has the potential to improve retrieval substantially.

\section{Conclusions}
\label{sec:conclusion}
We investigate whether the improved perplexity observed in \knn models can be translated into enhanced reasoning capabilities. We conduct extensive evaluation across 22 datasets. Our findings indicate that while \knn improve perplexity and can achieve better performance on memory-intensive tasks, they struggle with reasoning-intensive tasks, showing a disconnect between LM ability and task ability. Further qualitative analysis reveals that even when \knn produce correct answers, these are often the result of spurious correlations rather than actual reasoning. We believe this places an upper bound on the usefulness of these approaches compared to results from parametric models.

\section*{Limitations}
As we are limited by computing budget, we only build datastores up to 610 million tokens. It is unlikely although not impossible that larger datastores built on general web corpus like C4 will lead to better reasoning capabilities. Additionally, we only experiment with LLMs with seven- to eight-billion model parameters as the base models. The findings in this paper may not generalize to other, possibly larger, base models. 

\section*{Acknowledgements}
This work was supported by NSF IIS-1901030, NSF CAREER 2037519, and the IARPA HIATUS Program.  

\bibliography{custom}

\appendix
\section{More Implementation Details}
\label{sec:implementation}
Table~\ref{tab:task-oriented-datastore} presents the data sources of the Wiki datastore. Table~\ref{tab:hyperparameters} shows hyperparameters we use for different tasks.

\section{More Qualitative Analysis}
\label{sec:more-analysis}
We explain why retrieving from Math improves LMs on sentiment analysis. First, we consider a sentiment analysis example in Table~\ref{tab:qualitative-sentiment-explanation}. In this task, given a sentence, a model is required to predict whether the sentiment expressed is positive or negative. The sentence in the example expresses a positive sentiment; however, Llama-2 predicts the sentiment to be negative. \knn, when retrieving from Wiki, fail to find sentiment-related tokens, and hence also predict a negative sentiment. Performing retrieval from Math produced the correct sentiment. However, this is more coincidental rather than reflective of the model's capability, because, although the retrieved tokens display a positive sentiment, the retrieved contexts are not relevant to the test example. we observe that sentiment-related content is ubiquitous, regardless of the source we use to build the datastore. Even in math textbooks, we find many sentences that express sentiment.


\begin{table}[t]
    \centering
    \begin{tabular}{lrr}
    \toprule
        Corpus & Text Size & Tokens \\ \midrule
        Wikitext103 & 0.5GB & 140M \\
        Amazon & 0.07GB& 18M\\
        CC-NEWS & 1.6GB& 443M\\
        IMDB & 0.03GB& 8M\\
        Total & 2.2GB& 609M \\ \bottomrule
    \end{tabular}
    \caption{Statistics of each data source in the Wiki datastore.}
    \label{tab:task-oriented-datastore}
\end{table}

\begin{table}[t]
\centering
    \begin{tabular}{lccc}
    \toprule
    Data & $\lambda$ & $k$ & $\tau$ \\
    \midrule
    Llama2 + Wiki      & 0.2 & 2048 & 5.0 \\
   Llama3 + Wiki & 0.1 & 2048 & 5.0 \\
    Mistral + Wiki  & 0.1 &2048 & 10.0 \\
    \bottomrule
    \end{tabular}
    \hspace{2em}
    \begin{tabular}{lccc}
    \toprule
    Data & $\lambda$ & $k$ & $\tau$ \\
    \midrule
    Llama2 + Math   & 0.2 & 1600 & 5.0 \\
    Llama3 + Math & 0.1 & 2048 & 3.0 \\
    Mistral + Math & 0.1 & 2048 & 10.0 \\
    \bottomrule
    \end{tabular}
    \caption{Hyperparameters in $k$NN-LM. \textbf{Top}: Hyperparameters for Wiki datastore. \textbf{Bottom}: Hyperparameters for Math datastore .}
    \label{tab:hyperparameters}
\end{table}

\begin{table*}[ht]
    \centering
    \small
    \begin{tabular}{p {9cm} c c}
    \toprule
        Sentiment Example & Label & LM Pred  \\ \midrule 
            humorous, artsy, and even cute, in an off-kilter, dark, vaguely disturbing way. The sentence has a tone that is & Positive & Negative \\ \midrule
        Retrieved Context & Retrieved  & $k$NN-LM Pred \\
        \midrule
        \multicolumn{1}{c}{\textit{Wiki}} && \\
        $\bullet$ meta-commentator, Imhoff gives us a decidedly modern delivery. His speaking rhythms are staccato and his tone & bitter & \\
        $\bullet$ Collins, who has worked on more than 100 children books and won several awards: his tone is & fun & Negative \\ 
        $\bullet$ is her own narrator, so the thoughts and feelings of others are conveyed secondhand or are absent entirely. Her tone and language are at turns & honest & \\
        \multicolumn{1}{c}{\textit{Math}} && \\
        $\bullet$ preferred term is not ``Platonist'' but ``quasiëmpiricist'', a word Tymoczko lends a subtly & different & \\
        $\bullet$ ... or a horror film (group 2, $N_{H}=29$ ). The data are coded so that higher scores indicate a more & positive & \multirow{2}{*}{Positive} \\
        $\bullet$ the failure of the Intermediate Value Theorem is neither here nor there nor anywhere else to them. This is not a bad nor a & good & \\
        \midrule
    \end{tabular}
    \caption{A sentiment analysis example with predictions of the standard LM and \knn. We show tokens retrieved from each datastore and their proceeding tokens.}
    \label{tab:qualitative-sentiment-explanation}
\end{table*}

\begin{table*}[t]
    \centering
    \small
    \begin{tabular}{p{9cm} p{2.5cm} p{2.5cm}}
    \toprule
        HotpotQA Example & Label & LM Pred  \\ 
        \midrule
        who is older, Annie Morton or Terry Richardson? & \multirow{1}{2cm}{Terry Richardson} & \multirow{1}{2cm}{Terry Richardson} \\
        \\
        \midrule
        Retrieved context & Token & $k$NN-LM Pred\\
        \midrule
        $\bullet$ And she still wasn't done. Later she tweeted a warning to all women. ``My hard won advice: never get into an elevator alone with [Terry Gilliam.] Terry & \multirow{3}{*}{Gilliam} & \multirow{8}{*}{Terry Gilliam}\\
        $\bullet$ \#MeToo https://t.co/jPnFhfB5GQ - Ellen Barkin(@EllenBarkin) March 17, 2018Barkin got another shot in.
Terry & \multirow{2}{*}{Gilliam} & \\
        $\bullet$ I haven't posted about Christina Hendricks in a while but it's Valentine's Day and that makes me think of chocolate and chocolate reminds me of Christina Hendricks. And Christina & \multirow{3}{*}{Hend} & \\ \midrule
    \bottomrule
    \end{tabular}
    \caption{Another example from HotpotQA explains the impact of high-frequency proper nouns in the corpus on \knn predictions retrieving from Wikipedia.}
    \label{tab:qualitative-high_freq_2}
\end{table*}

\end{document}